\documentclass[lettersize,journal]{IEEEtran}
\usepackage{amsmath,amsfonts,amssymb}
\usepackage{algorithmic}
\usepackage{algorithm}
\usepackage{array}
\usepackage[caption=false,font=normalsize,labelfont=sf,textfont=sf]{subfig}
\usepackage{textcomp}
\usepackage{stfloats}
\usepackage{url}
\usepackage{verbatim}
\usepackage{graphicx}
\usepackage{cite}
\def\eg{\emph{e.g.}}
\def\ie{\emph{i.e.}}
\def\etal{\emph{et~al.}}
\usepackage{colortbl}
\usepackage[dvipsnames,svgnames,x11names]{xcolor}
\usepackage{booktabs}
\usepackage{multirow}
\usepackage{threeparttable}

\begin{document}

\title{Uncertainty-Aware Trajectory Prediction: \\ A Unified Framework Harnessing Positional and Semantic Uncertainties}

\author{Jintao~Sun,~Hu~Zhang,~Gangyi~Ding,~Zhedong~Zheng%
\IEEEcompsocitemizethanks{%
\IEEEcompsocthanksitem T.~Sun and G.~Ding are with the School of Computer Science and Technology, Beijing Institute of Technology, China 100081.
\protect\\ E-mail: 3120215524@bit.edu.cn, dgy@bit.edu.cn
\IEEEcompsocthanksitem H.~Zhang is with CSIRO DATA61, Australia. 
\protect\\ E-mail: hu1.zhang@csiro.au
\IEEEcompsocthanksitem Z.~Zheng is with the Faculty of Science and Technology and Institute of Collaborative Innovation, University of Macau, China 999078. 
\protect\\ E-mail: zhedongzheng@um.edu.mo
}
}

\markboth{Journal of \LaTeX\ Class Files,~Vol.~14, No.~8, August~2021}%
{Shell \MakeLowercase{\textit{et al.}}: A Sample Article Using IEEEtran.cls for IEEE Journals}


\maketitle

\begin{abstract}
Trajectory prediction seeks to forecast the future motion of dynamic entities, such as vehicles and pedestrians, given a temporal horizon of historical movement data and environmental context. A central challenge in this domain is the inherent uncertainty in real-time maps, arising from two primary sources: (1) positional inaccuracies due to sensor limitations or environmental occlusions, and (2) semantic errors stemming from misinterpretations of scene context. To address these challenges, we propose a novel unified framework that jointly models positional and semantic uncertainties and explicitly integrates them into the trajectory prediction pipeline. Our approach employs a dual-head architecture to independently estimate semantic and positional predictions in a dual-pass manner, deriving prediction variances as uncertainty indicators in an end-to-end fashion. These uncertainties are subsequently fused with the semantic and positional predictions to enhance the robustness of trajectory forecasts. We evaluate our uncertainty-aware framework on the nuScenes real-world driving dataset, conducting extensive experiments across four map estimation methods and two trajectory prediction baselines. Results verify that our method (1) effectively quantifies map uncertainties through both positional and semantic dimensions, and (2) consistently improves the performance of existing trajectory prediction models across multiple metrics, including minimum Average Displacement Error (minADE), minimum Final Displacement Error (minFDE), and Miss Rate (MR). Code will available at \url{https://github.com/JT-Sun/UATP}.
\end{abstract}


\section{Introduction}
\IEEEPARstart{A}{ccurate} and efficient prediction of future vehicle trajectories is critical for autonomous driving systems~\cite{9878832,Gu_Sun_Zhao_2021,ngiam2022scene,wu2023PPGeo}. To generate reliable trajectory predictions, autonomous vehicles must thoroughly understand and process their surrounding environment. High-definition (HD) maps are essential for this task. However, the dynamic nature of the environment presents significant challenges to accurate trajectory prediction. For instance, pedestrians sometimes suddenly enter a vehicle's path, weather and visibility conditions can fluctuate, obstacles obstruct the view, and sensor errors can introduce noise. These factors can cause discrepancies in the vehicle's perception of map information, thereby impacting trajectory prediction performance.

The existing trajectory prediction works concentrate on two key aspects. 
 \textbf{(1) One line of work focuses on the HD maps estimation.} 
The early works usually construct HD maps in an offline process, which heavily relies on simultaneous localization and mapping (SLAM) techniques~\cite{8594299,Zhang2014LOAMLO}. 
However, SLAM usually requires extra maintenance costs.
Therefore, some researchers resort to the bird's-eye view (BEV) representations~\cite{chen2022efficientrobust2dtobevrepresentation,Li_Wang_Li_Xie_Sima_Lu_Qiao_Dai_2022,9878966,9710288,MaskBEV,10.1145/3581783.3613798,10.1109/TPAMI.2024.3449912,liu2024monocular}, which uses deep neural networks to extract and fuse map information from multiple sensors and environmental data in an end-to-end manner. 
However, those methods typically do not provide a vectorized path, which represents the road as a sequence of interconnected keypoints. This representation allows for a more precise description of the road's geometric and topological characteristics. To further enhance the expressiveness of the map, 
some approaches~\cite{Li_Wang_Wang_Zhao_2022, pmlr-v202-liu23ax, MapTR,maptrv2,xu2024insmapperexploringinnerinstanceinformation,li2023lanesegnet}  have adopted a vectorized map format. This format not only preserves detailed environmental information but also aligns more closely with the structure of trajectory data, thereby facilitating downstream tasks such as path planning and trajectory prediction~\cite{yu2025clear,lin2024mmtec}.
\begin{figure*}[t]
   \begin{center}
    \centering{\includegraphics[width=\linewidth]{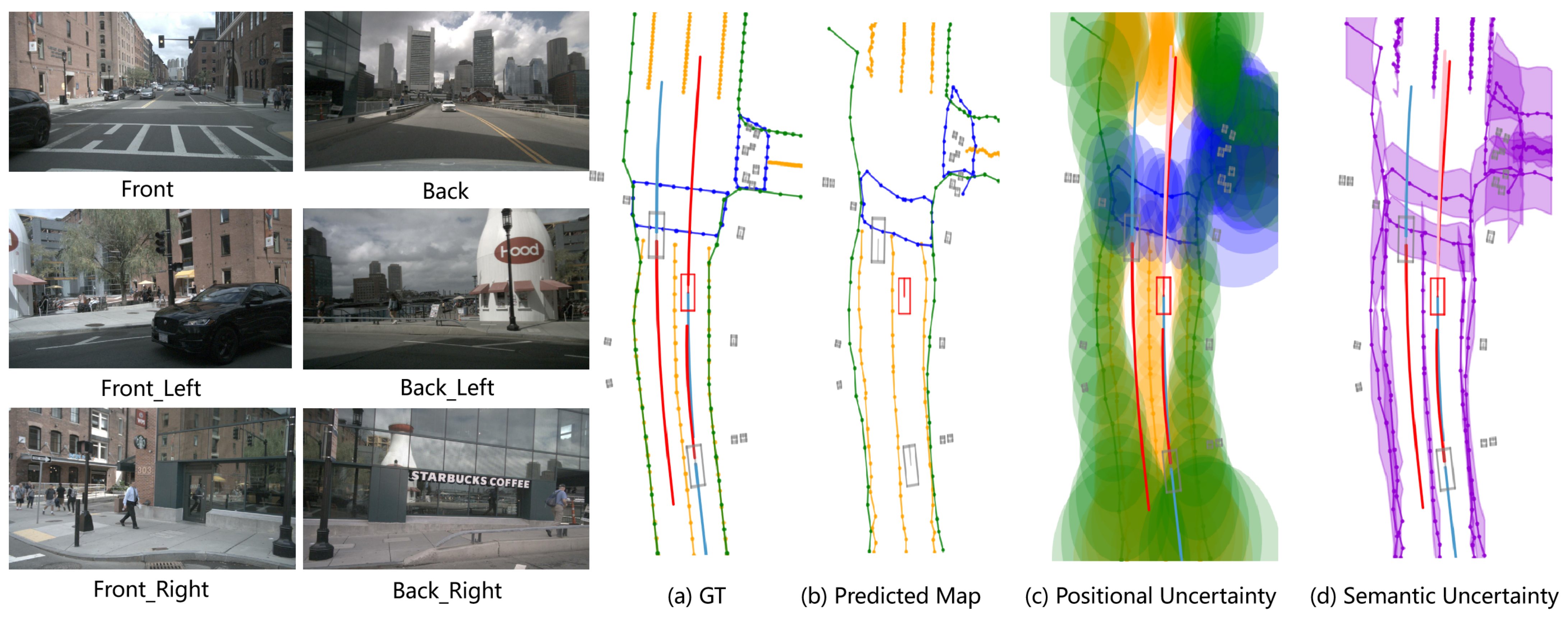}}

    \caption{
    \textbf{Motivation.} The $6$ images on the left are captured by $6$ different cameras on the vehicle. 
The map estimation remains challenging from RGB images, and thus inevitably contains noise, accumulating the error in the trajectory prediction.
Comparing the ground-truth HD map (a) and the predicted map in (b), we can see that the error usually occurs in uncertain areas. 
Therefore, in this work, we intend to leverage two types of uncertainty, \ie, positional uncertainty and semantic uncertainty, to indicate the map errors, mitigating the negative impacts. 
 (c) shows the positional uncertainty for three categories shown in three colors: \textcolor{ForestGreen}{green} for boundary, \textcolor{blue} {blue} for pedestrian crossing, \textcolor{YellowOrange}{orange} for divider, and \textcolor{red}{red} for the ego car. \textbf{The greater the positional uncertainty of the three categories, the larger the ellipse centered on the map element.}
 (d) shows the semantic uncertainty of our constructed high-definition map, where the \textcolor{Fuchsia}{purple} error band indicates the likelihood of an area being misclassified as another category.}
    \label{fig:1}
    \end{center}
\end{figure*}
\textbf{(2) Another line of work focuses on directly refining the trajectory prediction model.}
Some pioneering works~\cite{Cui_Radosavljevic_Chou_Lin_Nguyen_Huang_Schneider_Djuric_2019, Jain_Casas_Liao_Xiong_Feng_Segal_Urtasun_2019, Chai_Sapp_Bansal_Anguelov_2019, Liang_Jiang_Murphy_Yu_Hauptmann_2020} usually extract rasterized BEV features from image inputs via Convolutional Neural Networks (CNNs), 
while recent works apply transformers~\cite{Vaswani_Shazeer_Parmar_Uszkoreit_Jones_Gomez_Kaiser_Polosukhin_2017,9878832,li2023sttis} or GNNs~\cite{Gao_Sun_Zhao_Shen_Anguelov_Li_Schmid_2020, LaneGCN, Zeng_Liang_Liao_Urtasun_2021, 9795092, Zhao_Gao_Lan_Sun_Sapp_Varadarajan_Shen_Shen_Chai_Schmid_et_al,10.1145/3664647.3680792,li2025lanelevel} to capture the relationships within the vectorized map.
However, both lines of work suffer from the inherent data noise, such as occlusions, weather changes, and other environmental complexities (see Figure~\ref{fig:1} left), and have not explicitly conducted the noise modeling. 
As shown in Figure~\ref{fig:1}~(a) and (b), map estimation inevitably contains the noise. This leads to error accumulation during the trajectory prediction training process, ultimately affecting the final performance. 

Therefore, in this work, we intend to explicitly model noise during training and regularize the training process. 
\textbf{It is worth noting we do not remove the noise, but mitigate the negative impact of such noises}. Specifically, we account for uncertainty in predictions and illustrate the relationship between noise and uncertainty in Figure~\ref{fig:1}~(c)~and~(d). We observe that high noise in the input data leads to increased uncertainty in map estimation. To precisely characterize noise, we categorize it into two types: (1) noise causing positional errors, such as sensor inaccuracies or environmental occlusions (Figure~\ref{fig:1} (c)), and (2) noise leading to cognitive errors due to incorrect scene understanding (Figure~\ref{fig:1} (d)). These are modeled as positional uncertainty and semantic uncertainty, respectively. Implementation-wise, we introduce a dual-head structure where the primary head processes feature from ``res5c'' and the auxiliary head processes features from ``res4f'' of the ResNet50 backbone. ``res5c'' corresponds to the output of the final layer in the last ResNet50 block, while ``res4f'' refers to the output of the preceding block. Both heads regress semantic and positional information, with their differences serving as measures of uncertainty. For each type of prediction, semantic or positional, our model performs two independent regressions, and the variance between these predictions serves as an uncertainty indicator. The map elements, enriched with positional and semantic uncertainties, are then integrated into downstream trajectory prediction models, enabling the utilization of uncertainty context to enhance prediction accuracy. In summary, our contributions are as follows:
\begin{itemize}

\item We observe an inherent problem in map estimation for trajectory prediction, \ie, the presence of noise in HD maps. While it is impractical to eliminate this noise entirely, we propose a new approach that leverages two types of uncertainty, \ie, positional and semantic, to indicate and mitigate its negative impacts. By explicitly integrating these uncertainties as noise indicators into the model training process, our method effectively reduces the adverse effects of data noise, thereby enhancing the robustness and accuracy of trajectory predictions.
\item Albeit simple, our approach integrates seamlessly with four existing mapping estimation and two trajectory approaches, consistently improving prediction accuracy. For instance, when using MapTRv2-Centerline as the map backbone and HiVT as the trajectory prediction backbone, we boost minADE by 8\%, minFDE by 10\%, and MR by 22\% on the nuScenes dataset.
\end{itemize}

\section{Related Work}\label{related_work}
\textbf{Map-Informed Trajectory Prediction.}
Map-based trajectory forecasting is inherently tied to map estimation progress. Early methods employed rasterized HD map and agent representations processed via CNNs~\cite{Cui_Radosavljevic_Chou_Lin_Nguyen_Huang_Schneider_Djuric_2019, Jain_Casas_Liao_Xiong_Feng_Segal_Urtasun_2019, Chai_Sapp_Bansal_Anguelov_2019, Liang_Jiang_Murphy_Yu_Hauptmann_2020}. Vectorized approaches now dominate, processing raw polylines through two paradigms: GNN-based models~\cite{Gao_Sun_Zhao_Shen_Anguelov_Li_Schmid_2020, LaneGCN, Zeng_Liang_Liao_Urtasun_2021, Zhao_Gao_Lan_Sun_Sapp_Varadarajan_Shen_Shen_Chai_Schmid_et_al,10192373,10173747} that encode map-agent interactions via graph structures, and Transformer-based methods~\cite{Vaswani_Shazeer_Parmar_Uszkoreit_Jones_Gomez_Kaiser_Polosukhin_2017, Deo_Wolff_Beijbom_2021, Gu_Sun_Zhao_2021, Liu_Zhang_Fang_Jiang_Zhou_2021, 9878832, GuSongEtAl2024} leveraging cross-attention mechanisms. Specialized architectures include hierarchical Transformers~\cite{9878832}, topology-aware networks~\cite{li2023toponet}, and PV2BEV feature integration~\cite{10.1007/978-3-031-73004-7_24}. While Gu~\etal~\cite{GuSongEtAl2024} partially addresses map uncertainty via unified representations, existing methods still under-disentangle error sources. Orthogonal to architectural advances, detection-to-forecasting tightens the perception–prediction interface by conditioning on detector outputs (\eg, Zhang~\etal~\cite{10122126}); we make this interface uncertainty-aware by explicitly estimating semantic and positional uncertainties from the map and injecting them into predictors, which improves robustness under map noise.

\noindent\textbf{Online Map Estimation.}
Online map estimation dynamically constructs map representations using vehicle sensor data. Initial research primarily adopts rasterized BEV segmentation with CNNs~\cite{Chai_Sapp_Bansal_Anguelov_2019, Cui_Radosavljevic_Chou_Lin_Nguyen_Huang_Schneider_Djuric_2019, Liang_Jiang_Murphy_Yu_Hauptmann_2020, IntentNet}, while later works transition to vectorized paradigms using encoder-decoder architectures for direct polyline / polygon regression~\cite{Liu2022BEVFusionMM, Philion_Fidler_2020, Li_Wang_Li_Xie_Sima_Lu_Qiao_Dai_2022, Li_Wang_Wang_Zhao_2022, Dong2022SuperFusionML,IC-Mapper}. Subsequent advancements focus on efficiency and refinement: streamline point-set prediction~\cite{MapTR,maptrv2,xu2024insmapperexploringinnerinstanceinformation}, incorporate spatiotemporal attention~\cite{Yuan_2024_streammapnet}, introduce memory-based tracking~\cite{MapTracker},  enhance instance features via mask guidance~\cite{Liu_2024_CVPR}, and leverage distillation~\cite{MapDistill}. Despite progress, existing methods neglect inherent sensor and prediction noise. We propose uncertainty-aware modeling to enhance robustness.

\noindent\textbf{Uncertainty Learning.}
Deep learning uncertainty is generally categorized into two types: aleatoric uncertainty, which captures inherent observation noise, and epistemic uncertainty, which reflects model uncertainty due to limited data. Recent advances in uncertainty learning have proven essential for robust map estimation and trajectory forecasting under challenging conditions such as sensor noise and occlusions~\cite{qian2024unifieduq}. Prior works include LSTM-based uncertainty modeling for traffic prediction~\cite{Ma_Zhu_Zhang_Yang_Wang_Manocha_2019}, per-step motion uncertainty estimation~\cite{9878832}, segmentation for remote sensing images~\cite{10891590}, unsupervised object detection~\cite{9439889}, and generative approaches capturing behavioral diversity~\cite{Lv_Huang_Cao_2022}. 
Although these methods mark significant progress in coupling map uncertainty with prediction tasks~\cite{GuSongEtAl2024}, they often rely on oversimplified proxies for uncertainty. Motivated by these limitations, we propose a dual-level framework that explicitly disentangles positional uncertainty, addressing spatial inaccuracies from semantic uncertainty, which arises from class misidentification in map elements. This fine-grained decomposition empowers downstream predictors to dynamically prioritize more reliable map features, thereby significantly enhancing robustness in dynamic scenarios.

\section{Uncertainty Preliminary}
\subsection{Definitions}

1. \textbf{Model Structure:}
The deep learning model $ M $ includes a main head $ C_{\text{main}} $ and an auxiliary head $ C_{\text{aux}} $.
For an input sample $ x $, the prediction output of the main head is $ p_{\text{main}} = C_{\text{main}}(x) $, and the prediction output of the auxiliary head is $ p_{\text{aux}} = C_{\text{aux}}(x) $.

\noindent2. \textbf{Uncertainty:}
We focus on the model's \textit{Epistemic Uncertainty}, which is the uncertainty in the model parameters.
Assume the model parameters $ \theta $ are random variables with a prior distribution $ P(\theta) $.

\noindent3. \textbf{Prediction Difference:}
Define the prediction difference $ D(x) $ as:
\setcounter{equation}{0}
\begin{align}     
     D(x) = \| p_{\text{main}} - p_{\text{aux}} \|,
\end{align}     
where $ \| \cdot \| $ denotes a norm (\eg, L2 norm). 

\subsection{Mathematical Derivation}
\textbf{Model's Predictive Distribution.}
Assume the model's output is a probability distribution $ P(y|x, \theta) $, where $ y $ is the class label, $ x $ is the input sample, and $ \theta $ are the model parameters.

\noindent\textbf{Posterior Predictive Distribution.}
According to Bayes' theorem, the posterior predictive distribution of the model can be expressed as:
\begin{align}
P(y|x) = \int P(y|x, \theta) P(\theta|x) d\theta,
\end{align}
where $ P(\theta|x) $ is the posterior distribution of the model parameters.

\noindent\textbf{Parameter Uncertainty.} The uncertainty in the parameters can be measured by the variance of the posterior distribution:
\begin{align}
\text{Var}(\theta|x) = \mathbb{E}_{\theta|x}[(\theta - \mathbb{E}_{\theta|x}[\theta])^2] = \mathbb{E}_{\theta|x}[\theta^2] - (\mathbb{E}_{\theta|x}[\theta])^2.
\end{align}

\noindent\textbf{Prediction Difference and Parameter Uncertainty.}
To relate the prediction difference $ D(x) $ to parameter uncertainty, we need to consider the predictions of the main and auxiliary heads. Assume the parameters of the main head and auxiliary head are $ \theta_{\text{main}} $ and $ \theta_{\text{aux}} $, respectively, and they have the same prior distribution, \ie, $ P(\theta_{\text{main}}) = P(\theta_{\text{aux}}) $. The predictions of the main and auxiliary heads can be expressed as:
\begin{align}
p_{\text{main}} = \mathbb{E}_{\theta_{\text{main}}|x}[P(y|x, \theta_{\text{main}})].
\end{align}
\begin{align}
p_{\text{aux}} = \mathbb{E}_{\theta_{\text{aux}}|x}[P(y|x, \theta_{\text{aux}})].
\end{align}

\noindent\textbf{Expression for Prediction Difference.} Assume the difference in predictions can be approximated by First-order Taylor Expansion:
\begin{align}
p_{\text{main}} - p_{\text{aux}} \approx \mathbb{E}_{\theta|x}[\nabla_{\theta} P(y|x, \theta) \cdot (\theta_{\text{main}} - \theta_{\text{aux}})],
\end{align}
where $ \nabla_{\theta} P(y|x, \theta) $ is the gradient of $ P(y|x, \theta) $ with respect to $ \theta $. Thus, the prediction difference $ D(x) $ can be expressed as:
\begin{align}
D(x) = \| p_{\text{main}} - p_{\text{aux}} \| \approx \| \mathbb{E}_{\theta|x}[\nabla_{\theta} P(y|x, \theta) \cdot (\theta_{\text{main}} - \theta_{\text{aux}})] \|
\end{align}

\noindent\textbf{Relationship Between Prediction Difference and Parameter Uncertainty.} To simplify the analysis, assume $ \theta_{\text{main}} $ and $ \theta_{\text{aux}} $ are independently and identically distributed (i.i.d.). Then:
\begin{equation}
\resizebox{\columnwidth}{!}{%
  $\displaystyle
  \begin{split}
  \mathbb{E}_{\theta\mid x}\bigl[\|\nabla_{\theta}P(y\mid x,\theta)\cdot(\theta_{\mathrm{main}}-\theta_{\mathrm{aux}})\|^2\bigr]
  \approx{}&\mathbb{E}_{\theta\mid x}\bigl[\|\nabla_{\theta}P(y\mid x,\theta)\|^2\bigr]\\
  &\quad\cdot\mathbb{E}_{\theta\mid x}\bigl[(\theta_{\mathrm{main}}-\theta_{\mathrm{aux}})^2\bigr].
  \end{split}
  $%
}
\end{equation}

By noting that
$\displaystyle
  \mathbb{E}_{\theta\mid x}[(\theta_{\mathrm{main}}-\theta_{\mathrm{aux}})^2]
  =\,\mathrm{Var}(\theta\mid x),
$
we therefore have:
\begin{equation}
\resizebox{\columnwidth}{!}{%
  $\displaystyle
    \mathbb{E}_{\theta\mid x}\bigl[\|\nabla_{\theta}P(y\mid x,\theta)\cdot(\theta_{\mathrm{main}}-\theta_{\mathrm{aux}})\|^2\bigr]
    \approx \,\mathbb{E}_{\theta\mid x}\bigl[\|\nabla_{\theta}P(y\mid x,\theta)\|^2\bigr]\,
    \mathrm{Var}(\theta\mid x)
  $%
}
\end{equation}

Assuming $ k = \mathbb{E}_{\theta|x}[\| \nabla_{\theta} P(y|x, \theta) \|^2] $, which is positive, we get:
\begin{align}
\mathbb{E}_{\theta|x}[\| \nabla_{\theta} P(y|x, \theta) \cdot (\theta_{\text{main}} - \theta_{\text{aux}}) \|^2] \approx k \cdot \text{Var}(\theta|x)
\end{align}

Thus, the prediction difference $ D(x) $ can be expressed as:
\begin{align}
D(x) \approx \sqrt{k \cdot \text{Var}(\theta|x)}
\end{align}

Simplifying further, we obtain:
\begin{align}
D(x) \propto \sqrt{\text{Var}(\theta|x)}
\end{align}
\subsection{Mathematical Conclusion}

From the above derivation, we have shown that the prediction difference $ D(x) $ is proportional to the square root of the model's uncertainty $ \sqrt{\text{Var}(\theta|x)} $. Therefore, the prediction difference $ D(x) $ can serve as a measure of the model's uncertainty for a given sample.
\begin{equation*}
\boxed{%
  \parbox{0.95\columnwidth}{\centering
    The prediction difference $D(x)$ is proportional to the square root of the model’s uncertainty\\
    $\sqrt{\mathrm{Var}(\theta\mid x)}$.
  }%
}
\end{equation*}

\noindent\textbf{Discussion. Uncertainty Comparison between existing works and ours.} In this discussion, we mainly compare the typical uncertainty quantification approach proposed by Gu \etal~\cite{GuSongEtAl2024} with our method. Gu~\etal~define uncertainty using class probability, denoted as $P(y|x, \theta)$, which represents the probability that an input $x$ belongs to class $y$ given model parameters $\theta$. In their framework, this class probability directly serves as the uncertainty measure. In contrast, our approach employs two independent prediction heads: a main head parameterized by $\theta_{\text{main}}$ and an auxiliary head parameterized by $\theta_{\text{aux}}$. We quantify uncertainty through the prediction difference, defined as $D(x) = \| p_{\text{main}} - p_{\text{aux}} \|$, which captures variability arising from differences in the model parameters.

The key distinction between these methods lies in their handling of parameter uncertainty. For a fixed $\theta$, the class probability $P(y|x, \theta)$ in Gu~\etal's approach remains deterministic and does not account for variability in $\theta$. Conversely, our prediction difference $D(x)$ is proportional to $\sqrt{\text{Var}(\theta|x)}$, directly reflecting the sensitivity of predictions to changes in model parameters. A larger $D(x)$ indicates greater variance in $\theta$, thereby signaling higher uncertainty. This makes our approach particularly effective in capturing parameter uncertainty, offering a more robust measure compared to the static class probability used by Gu~\etal

\section{Method}
\subsection{Uncertainty Estimation}~\label{sec:3.1}
\begin{figure*}[t]
    \centering
    \includegraphics[width=\linewidth]{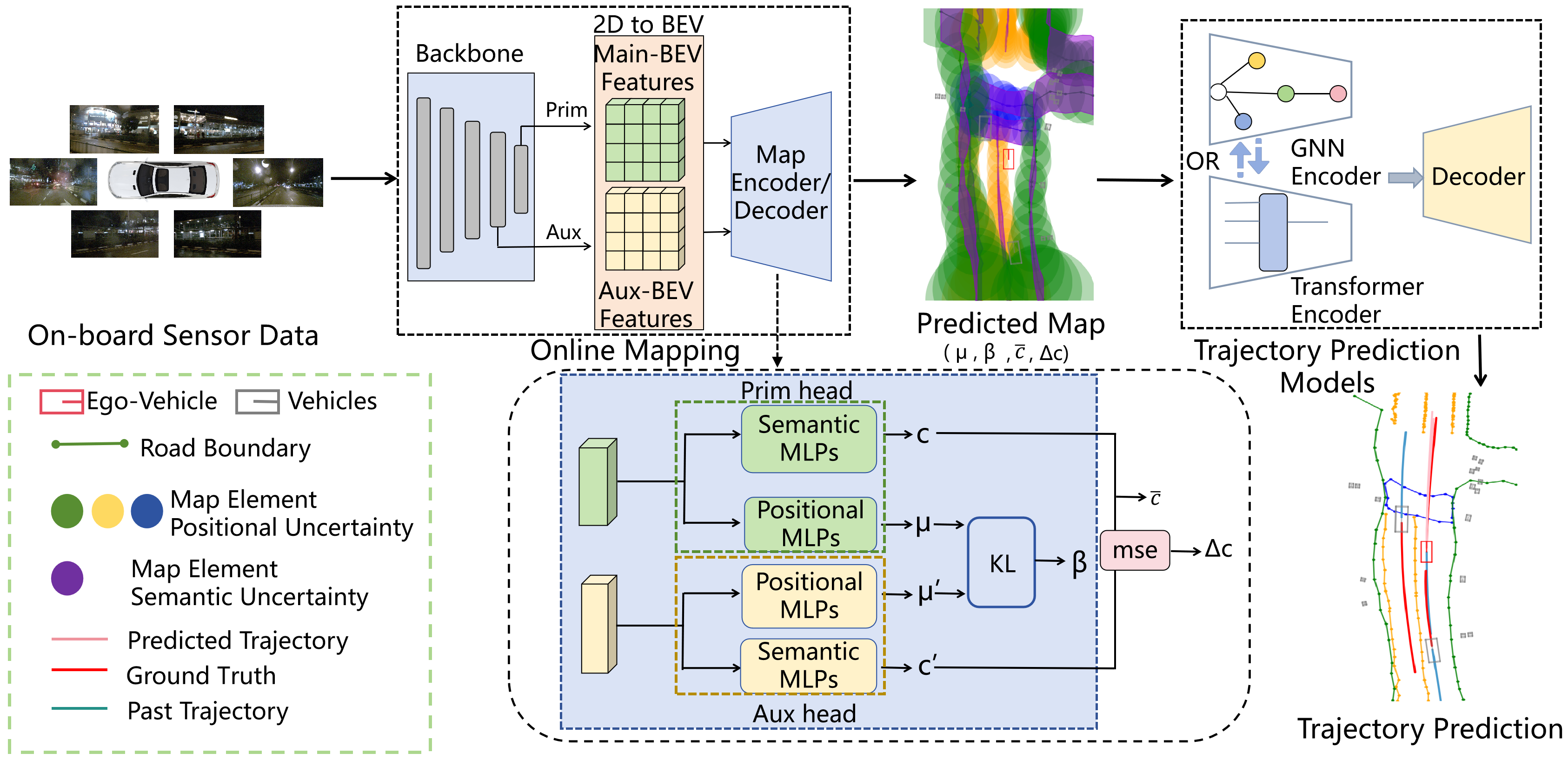}
    \caption{
    \textbf{Overall pipeline.} 
    Firstly, given six images from the vehicle camera, we extract 2D features via a visual backbone and transform them into BEV features. Specifically, we harness the primary and auxiliary heads to predict two BEV features given different-level feature maps.
    Secondly, during the map estimation stage (bottom), we perform location regression and semantic regression on the features from both primary and auxiliary heads. 
    The resulting primary and auxiliary map element vectors ($\boldsymbol{\mu}$ and $\boldsymbol{\mu'}$ ) are used to calculate KL divergence as positional uncertainty $\boldsymbol{\beta}$. 
    Similarly, we obtain the semantic scores $\boldsymbol{c}$ and $\boldsymbol{c'}$ from the corresponding MLPs, and then we could derive the semantic uncertainty $\boldsymbol{\Delta c}$. 
    Thirdly, we concatenate the high-definition map location information $\boldsymbol{\mu}$, the mean semantic score $\boldsymbol{\bar{c}}$, and their uncertainties $(\boldsymbol{\beta, \Delta c})$, as the input of the downstream model (GNN or Transformer Encoder) to facilitate the scene understanding for trajectory prediction.
    }
    \label{fig:arch}
\end{figure*}

In Figure~\ref{fig:arch}, we present an overview of our trajectory prediction pipeline. We start by extracting 2D features from vehicle camera images and transforming them into BEV features. To capture both positional and semantic uncertainty, we implement a dual-head architecture with primary and auxiliary heads of identical structure. Each BEV feature is processed by both heads, which independently generate positional and semantic predictions. We then perform location and semantic regression on the outputs from both heads. Positional uncertainty is quantified by computing the KL divergence between the primary and auxiliary location predictions. For semantic information, we calculate the mean and MSE of the classification probabilities from both heads to obtain the average semantic score and its uncertainty. Finally, the high-definition map location data, semantic information, and their corresponding uncertainties are integrated into the encoded map representation used by the downstream trajectory prediction model. The following sections provide further details on each pipeline component.

\noindent\textbf{Positional Uncertainty.} 
In particular, to estimate the position of map elements, we first adopt an MLP-based structure to regress a two-dimensional vector representing the normalized BEV coordinates $(x, y)$ of each map element. We then design an auxiliary head with a structure similar to the primary head. The only difference is that we additionally introduce one dropout layer to increase the variability in prediction. Thus, for each map element, we obtain the primary map element vector $\boldsymbol{\mu}$ and the auxiliary map element vector $\boldsymbol{\mu}' $. Following Gu~\etal~\cite{GuSongEtAl2024}, we apply the Laplace distribution to both $\boldsymbol{\mu}$ and $\boldsymbol{\mu}' $. To better estimate positional uncertainty, we calculate the KL divergence between $\boldsymbol{\mu}$ and $\boldsymbol{\mu}' $ and apply that to quantify positional uncertainty for each map element. Mathematically, this process is defined as:
\begin{align}
\boldsymbol{\beta} =\mathbb{E} \left [ \boldsymbol{\mu} \log\left ( \frac{\boldsymbol{\mu}}{\boldsymbol{\mu}' }  \right )    \right ].
\end{align}
A significant divergence between the predicted vectors from the two regression heads results in a large approximate variance, reflecting increased model uncertainty. This uncertainty provides a more detailed characterization of positional noise for each map element and captures the confidence of the model in its predictions.

\noindent\textbf{Semantic Uncertainty.}
For semantic uncertainty, we also utilize two heads to process BEV features from the input, each independently producing a set of class scores. We denote the classification probability from primary and auxiliary heads as $\boldsymbol{c}$ and $\boldsymbol{c}'$, respectively. For better usage in downstream tasks, we calculate the mean of $\boldsymbol{c}$ and $\boldsymbol{c}'$, denoted as $\boldsymbol{\bar{c}} = \frac{1}{2}(\boldsymbol{c}+\boldsymbol{c}')$, to serve as the updated confidence score for the map element. Meanwhile, we compute the MSE between the confidence scores $\boldsymbol{c}$ and $\boldsymbol{c}'$ from the primary and auxiliary heads, using this divergence as a supplementary uncertainty measure $\boldsymbol{\Delta c}$ for semantic classification confidence: 
\begin{align}
\boldsymbol{\Delta c} = (\boldsymbol{c} - \boldsymbol{c}')^2 .
\end{align}
Our semantic uncertainty for map elements comprises two key components: the average classification confidence score and the supplementary uncertainty derived from the MSE between the classification probabilities of the two heads.

\noindent\textbf{Discussion.} \textbf{(1) Why use an auxiliary head to estimate uncertainty?}
By introducing an auxiliary head that extracts features from RGB images, the model captures a different receptive field compared to the primary head. While the primary head focuses on deeper-level features, the auxiliary head processes relatively shallower ones. This multi-layered feature extraction ensures that both deep and shallow image features are considered. The variation in feature extraction between the two heads provides valuable insights for uncertainty estimation. Discrepancies in predictions from the main and auxiliary heads help gauge the level of uncertainty. Additionally, when estimating both positional and semantic uncertainty, we introduce a dropout layer after the auxiliary head. This introduces variability in the positional and semantic features during training, amplifying the differences between the predictions. These enhanced discrepancies boost the ability of the model to estimate uncertainty, thereby enhancing the robustness and accuracy of the trajectory predictions.

\noindent\textbf{(2) Why perform positional and semantic uncertainty separately?} The core objective is to enrich map elements with more diverse and accurate information, while simulating real-world conditions such as occlusions and sensor errors, which can affect map prediction accuracy. These factors can lead to imprecise location predictions, resulting in errors in subsequent agent trajectory predictions. Additionally, downstream tasks rely on map elements that contain both positional and semantic information. By separately estimating positional and semantic uncertainties, we provide a more comprehensive representation of the environment. This allows downstream prediction networks to better leverage both spatial positions and their corresponding semantic features, leading to more reliable and robust trajectory predictions.

\noindent\textbf{(3) The compatibility of the proposed uncertainty.}
Our uncertainty estimation method is highly compatible with advanced map element estimation approaches. We verify this by integrating our uncertainty estimation into four state-of-the-art online HD mapping methods: MapTR~\cite{MapTR}, MapTRv2~\cite{maptrv2}, MapTRv2-Centerline, and StreamMapNet~\cite{Yuan_2024_streammapnet}.
Both MapTR~\cite{MapTR} and MapTRv2~\cite{maptrv2} utilize an encoder-decoder architecture to transform RGB images into BEV features using the LSS method~\cite{PhilionFidler2020LSS}. When incorporating our proposed uncertainty, we adopt a perception processing method similar to prior work~\cite{GuSongEtAl2024}. This ensures that the four types of map element information generated by these models are constrained within a perception range centered around the autonomous vehicle, with a longitudinal range of 60 meters and a lateral range of 30 meters. This enriched and uncertainty-aware map information enhances the accuracy and robustness of trajectory prediction learning. By providing more comprehensive and reliable map data, our approach enables downstream models to better handle real-world conditions and uncertainties, leading to improved performance in trajectory prediction tasks.

\subsection{Uncertainty-aware Trajectory Prediction}

\begin{table*}[t]
\centering
\setlength{\tabcolsep}{6pt}
\renewcommand{\arraystretch}{1.2}
\caption{Quantitative results of eight experiments combining 4 high-definition map estimation models and $2$ trajectory prediction models on the nuScenes~\cite{Caesar_Bankiti_Lang_Vora_Liong_Xu_Krishnan_Pan_Baldan_Beijbom_2020} dataset are presented. Overall, we observe that our method, which integrates both positional and semantic uncertainties, outperforms previous approaches in enhancing the prediction model performance, with the most significant improvement seen in the MapTRv2-centerline method and StreamMapNet method.}
\label{tbl:result} 
\scalebox{0.90}{
\begin{tabular}{l lll lll}
\toprule 
Online HD Map Method & \multicolumn{6}{|c}{Trajectory Prediction Method}  \\
\cmidrule(lr){2-7}
& \multicolumn{3}{|c|}{HiVT~\cite{9878832}}                                                & \multicolumn{3}{c}{DenseTNT~\cite{Gu_Sun_Zhao_2021}}                                          \\ 
 \cmidrule(lr){2-7} 
& \multicolumn{1}{|c}{minADE $\downarrow$}  & minFDE $\downarrow $  &\multicolumn{1}{l|}{ MR $\downarrow $\phantom{a} }                      & minADE $\downarrow $                & minFDE $\downarrow$               & MR $\downarrow $                   \\ 
\midrule %
\multicolumn{1}{l|}{MapTR~\cite{MapTR}}                                                        & 0.4015                 & 0.8418                 & \multicolumn{1}{l|}{0.0981}                 & 1.091                  & 2.058                & 0.3543                \\
\multicolumn{1}{l|}{MapTR~\cite{MapTR} +~\cite{GuSongEtAl2024} }             & 0.3854                 & 0.7909                 & \multicolumn{1}{l|}{0.0834}                 & 1.089                  & 2.006                & 0.3499                \\
\multicolumn{1}{l|}{MapTR~\cite{MapTR} + Ours}                                                 & \textbf{0.3660 \color{ForestGreen}{$(-5\%)$}}  & \textbf{0.7564 \color{ForestGreen}{$(-5\%)$}} & \multicolumn{1}{l|}{\textbf{0.0745 \color{ForestGreen}{$(-11\%)$}}} & \textbf{0.954 \color{ForestGreen}{$(-13\%)$}}  & \textbf{1.909 \color{ForestGreen}{$(-5\%)$}} & \textbf{0.3429 \color{ForestGreen}{$(-2\%)$}} \\ 
\midrule
\multicolumn{1}{l|}{MapTRv2~\cite{maptrv2}}                                                      & 0.4057                 & 0.8499                 & \multicolumn{1}{l|}{0.0992}                 & 1.214                  & 2.312                & 0.4138                \\
\multicolumn{1}{l|}{MapTRv2~\cite{maptrv2} +~\cite{GuSongEtAl2024} }           & 0.3930                 & 0.8127                 & \multicolumn{1}{l|}{0.0857}                 & 1.262                  & 2.340                & \textbf{0.3912}       \\
\multicolumn{1}{l|}{MapTRv2~\cite{maptrv2} + Ours }                                              & \textbf{0.3697 \color{ForestGreen}{$(-3\%)$}}  & \textbf{0.7621 \color{ForestGreen}{$(-6\%)$}} & \multicolumn{1}{l|}{\textbf{0.0787 \color{ForestGreen}{$(-8\%)$}}} & \textbf{1.099 \color{ForestGreen}{$(-13\%)$}}  & \textbf{2.235 \color{ForestGreen}{$(-5\%)$}} & 0.4230 \color{Peach}{$(+8\%)$}          \\ 
\midrule
\multicolumn{1}{l|}{MapTRv2-Centerline~\cite{maptrv2}}                                           & 0.3790                 & 0.7822                 & \multicolumn{1}{l|}{0.0853}                 & 0.8466                 & 1.345                & 0.1520                \\
\multicolumn{1}{l|}{MapTRv2-Centerline~\cite{maptrv2} +~\cite{GuSongEtAl2024}} & 0.3727                 & 0.7492                 & \multicolumn{1}{l|}{0.0726}                 & 0.8135                 & \textbf{1.311}       & 0.1593                \\
\multicolumn{1}{l|}{MapTRv2-Centerline~\cite{maptrv2} + Ours}                                    & \textbf{0.3427 \color{ForestGreen}{$(-8\%)$}} & \textbf{0.6763 \color{ForestGreen}{$(-10\%)$}} & \multicolumn{1}{l|}{\textbf{0.0570 \color{ForestGreen}{$(-22\%)$}}} & \textbf{0.7419 \color{ForestGreen}{$(-9\%)$}} & 1.341 \color{Peach}{$(+2\%)$}           & \textbf{0.1506 \color{ForestGreen}{$(-6\%)$}} \\ 
\midrule
\multicolumn{1}{l|}{StreamMapNet~\cite{Yuan_2024_streammapnet}}                                                 & 0.3972                 & 0.8186                 & \multicolumn{1}{l|}{0.0926}                 & 0.9492                 & 1.740                & 0.2569                \\
\multicolumn{1}{l|}{StreamMapNet~\cite{Yuan_2024_streammapnet} +~\cite{GuSongEtAl2024}}       & 0.3848                 & 0.7954                 & \multicolumn{1}{l|}{0.0861}                 & 0.9036        & 1.645       & \textbf{0.2359}       \\
\multicolumn{1}{l|}{StreamMapNet~\cite{Yuan_2024_streammapnet} + Ours }                                        & \textbf{0.3711 \color{ForestGreen}{$(-7\%)$}}  & \textbf{0.7745 \color{ForestGreen}{$(-10\%)$}}  & \multicolumn{1}{l|}{\textbf{0.0796 \color{ForestGreen}{$(-22\%)$}}} & \textbf{0.8065} \color{ForestGreen}{$(-11\%)$}   & \textbf{1.600 } \color{ForestGreen}{$(-3\%)$} & 0.2418 \color{Peach}{$(+2\%)$}
      \\ 
\bottomrule 
\end{tabular}
}
\end{table*}

Trajectory prediction aims to forecast the future trajectories of agents in highly dynamic environments. Typically, vertex coordinates are encoded using multilayer perceptrons (MLPs) within the encoder, followed by the integration of GNNs or Transformer-based attention layers to capture long-term dependencies among entities.

Our uncertainty-aware trajectory prediction method specifically incorporates the positional uncertainty and semantic uncertainty introduced in Section~\ref{sec:3.1} during the encoder process. Our input for the trajectory prediction consists of four types of uncertainty information: map positional uncertainty $\boldsymbol{\mu}$, differentiable information $\boldsymbol{\beta}$, semantic class probability 
$\boldsymbol{\bar{c}}$ derived from semantic uncertainty estimation, and supplementary semantic variation 
$\boldsymbol{\Delta c}$. We combine these four uncertainty representations into a unified encoding and form the uncertainty-aware map information. This process can be formulated as: $\mathrm{MLPs}\left [\mathrm{concat }\left ( \boldsymbol{\mu} ,\boldsymbol{\beta },\boldsymbol{\bar{c}},\boldsymbol{\Delta c}  \right )   \right ],$
where $\mathrm{concat}$ denotes the concatenation operation; $\boldsymbol{\bar{c}}, \boldsymbol{\Delta c} \in \Phi ^{C-1}$ represent the probability simplex with $C$ classes.

Our uncertainty-aware trajectory prediction method seamlessly integrates with two state-of-the-art models: HiVT~\cite{9878832} and DenseTNT~\cite{Gu_Sun_Zhao_2021}. HiVT, a Transformer-based approach, treats vectorized map elements as a sequence of tokens. In our framework, map elements augmented with positional and semantic uncertainty are input as point sets into the HiVT encoder, where these uncertainties are concatenated and jointly encoded during the local encoding stage. Conversely, DenseTNT, a GNN-based model, directly encodes map elements with uncertainty information using VectorNet~\cite{pmlr-v202-liu23ax}.

\noindent\textbf{Discussion. What are the advantages of the proposed uncertainty-aware trajectory prediction method?} Accurate vehicle trajectory prediction is highly dependent on HD map data, as map elements are crucial for predicting agent trajectories. While some previous methods~\cite{GuSongEtAl2024} have utilized map uncertainty to enhance trajectory predictions, they often focus solely on Laplace-distributed location uncertainties and provide only basic class probabilities. Different from existing works, our proposed approach incorporates both positional and semantic uncertainties, thereby enriching the map elements with more comprehensive uncertainty information. This enhanced representation allows the prediction model to better leverage contextual information, leading to more accurate and robust trajectory forecasting.

\section{Experiment}
\subsection{Experiment Setup}

\begin{table}[t]
\centering
\setlength{\tabcolsep}{3pt}
\renewcommand{\arraystretch}{1.2}
\caption{Ablation study on our main components, \ie, positional uncertainty and semantic uncertainty.
 Unc\_pos denotes the positional uncertainty method, while Unc\_sem represents the semantic uncertainty method. We use a checkmark $\checkmark$ to indicate whether the method is applied. * means part of our uncertainty.}
\label{tbl:ablation} 
\scalebox{0.85}{
\begin{tabular}{l|l|c|c|lll}
\toprule %
Online HD       &  Trajectory      & Unc & Unc & minADE $\downarrow$ & minFDE $\downarrow$ & MR $\downarrow$    \\ 
Map Method & Prediction Method & pos & sem & & &  \\
\midrule %
MapTR~\cite{MapTR} & HiVT~\cite{9878832}         &               &               & 0.4015 & 0.8418 & 0.0981 \\
Baseline~\cite{GuSongEtAl2024} & HiVT~\cite{9878832}    &             &               & 0.3854 & 0.7909 & 0.0834 \\
Ours* & HiVT~\cite{9878832}         & \checkmark             &               & 0.3717 & 0.7820 & 0.0829 \\
Ours* & HiVT~\cite{9878832}       &               & \checkmark             & \textbf{0.3643} & 0.7573 & 0.0812 \\
\rowcolor{gray!25}
Ours & HiVT~\cite{9878832}        & \checkmark             & \checkmark             & 0.3660 & \textbf{0.7564} & \textbf{0.0745} \\ 
\midrule %
MapTR~\cite{MapTR} & DenseTNT~\cite{Gu_Sun_Zhao_2021}    &               &               & 1.091  & 2.058  & 0.3543 \\
Baseline~\cite{GuSongEtAl2024} & DenseTNT~\cite{Gu_Sun_Zhao_2021}    &             &               & 1.089 & 2.006 & 0.3499 \\
Ours* & DenseTNT~\cite{Gu_Sun_Zhao_2021}     & \checkmark             &               & 1.093  & 2.207 & 0.4286 \\
Ours* & DenseTNT~\cite{Gu_Sun_Zhao_2021}     &               & \checkmark             & 0.987 & 1.935 & 0.3456 \\
\rowcolor{gray!20}
Ours & DenseTNT~\cite{Gu_Sun_Zhao_2021}     & \checkmark            & \checkmark             & \textbf{0.954}  & \textbf{1.909}  & \textbf{0.3429} \\ 
\bottomrule %
\end{tabular}
}
\end{table}

\noindent\textbf{Dataset.} 
We evaluate our method using the widely recognized large-scale nuScenes dataset~\cite{Caesar_Bankiti_Lang_Vora_Liong_Xu_Krishnan_Pan_Baldan_Beijbom_2020}, comprising 1,000 diverse driving scenes split into 500 for training, 200 for validation, and 150 for testing. Each scene spans approximately 20 seconds, featuring RGB images captured by six synchronized cameras that collectively provide a comprehensive 360° horizontal field-of-view around the ego-vehicle. Sensor data is recorded at 10 Hz, with keyframe annotations available at 2 Hz. Additionally, the dataset provides accurate ground-truth HD maps, multi-sensor inputs, and precisely tracked agent trajectories.
Our approach employs the unified trajdata interface~\cite{ivanovic2023trajdata}, following Gu~\etal~\cite{GuSongEtAl2024}, to standardize data handling between vectorized map estimation and downstream trajectory prediction models. To ensure seamless compatibility among diverse prediction and mapping models, we utilize the trajdata temporal interpolation utility~\cite{ivanovic2023trajdata}, as introduced by Gu~\etal~\cite{GuSongEtAl2024}. This utility upscales nuScenes trajectory data frequency from 2 Hz to 10 Hz, enabling frequency alignment. Subsequently, each prediction model forecasts vehicle motion three seconds into the future based on two seconds of preceding vehicle trajectories.

\noindent\textbf{Metrics.}
For evaluating trajectory prediction performance, we adopt three standard metrics commonly utilized in recent prediction benchmarks: minimum Average Displacement Error (minADE), minimum Final Displacement Error (minFDE), and Miss Rate (MR). Specifically, for each agent generating six trajectory predictions, minADE quantifies the average Euclidean distance (in meters) between the most accurate predicted trajectory and the corresponding ground truth over the entire prediction horizon. minFDE assesses the Euclidean error between the final predicted position and the ground truth endpoint, identifying the best prediction as the trajectory with the minimal endpoint deviation. Lastly, MR measures the proportion of trajectory predictions whose endpoints exceed a threshold of two meters from the ground truth, providing insight into prediction reliability.

\noindent\textbf{Implementation Details.}
All models are trained on four NVIDIA GeForce RTX A6000 GPUs, each with 49 GB of memory. We use four independent methods and adjust network structures to account for positional and semantic uncertainty, resulting in slight parameter changes compared to the baseline. Additionally, due to structural differences, we apply separate hyperparameter settings for each. For a fair comparison, we follow the hyper-parameter setting in~\cite{GuSongEtAl2024} as shown in the Table~\ref{tab:training_hyperparameters}. For all map estimation models, we set the learning rate to $1.0\times10^{-4}$, regression loss weight to 0.03, and gradient norm to 3. 
Similarly, for the two downstream trajectory prediction models, the model input information changes and the model structures differ. Therefore, we use different hyperparameters for training each model, as shown in Table~\ref{tab:hyperparameters_trajectory}. We set the learning rate to $3.5\times10^{-4}$ for all four map prediction models with the trajectory prediction model HiVT. Four different learning rates from $2.5\times10^{-3}$ to $3.5\times10^{-3}$ are set for different map prediction models with the trajectory prediction model DenseTNT. When using the HiVT model, the batch size is set to 32. For DenseTNT, the batch size is set to 16. The dropout rate for all trajectory prediction models is 0.1. All other hyperparameters in these two trajectory prediction models are unchanged.

\begin{table}[t]
    \centering
    \caption{Map prediction training hyperparameters.}
    \scalebox{0.75}{
    \begin{tabular}{l|c c c}
        \toprule
        \textbf{Method} & \textbf{Regression Loss Weight} & \textbf{LR} & \textbf{Gradient Norm} \\ 
        \midrule
        MapTR~\cite{MapTR} & 0.03 & 1.0E-4 & 3 \\
        MapTRv2~\cite{maptrv2} & 0.03 & 1.0E-4 & 3 \\
        MapTRv2-Centerline~\cite{maptrv2} & 0.03 & 1.0E-4 & 3 \\
        StreamMapNet~\cite{Yuan_2024_streammapnet} & 2.00 & 1.0E-4 & 3 \\
        \bottomrule
    \end{tabular}
    }
    \label{tab:training_hyperparameters}
\end{table}

\begin{table}[t]
    \centering
    \caption{Hyperparameters chosen for different trajectory prediction methods.}
    \scalebox{0.8}{
    \begin{tabular}{l|c c c}
        \toprule
        \textbf{Online HD Map Method} & \textbf{LR} & \textbf{Batch Size} & \textbf{Dropout} \\ 
        \midrule
        MapTR~\cite{MapTR} + HiVT~\cite{9878832} & 3.5E-4 & 32 & 0.1 \\
        MapTR~\cite{MapTR} + DenseTN~\cite{Gu_Sun_Zhao_2021} & 3.0E-3 & 16 & 0.1 \\
        \midrule
        MapTRv2~\cite{maptrv2} + HiVT~\cite{9878832} & 3.5E-4 & 32 & 0.1 \\
        MapTRv2~\cite{maptrv2} + DenseTNT~\cite{Gu_Sun_Zhao_2021} & 2.0E-3 & 16 & 0.1 \\
        \midrule
        MapTRv2-Centerline~\cite{maptrv2} + HiVT~\cite{9878832} & 3.5E-4 & 32 & 0.1 \\
        MapTRv2-Centerline~\cite{maptrv2} + DenseTNT~\cite{Gu_Sun_Zhao_2021} & 3.5E-3 & 16 & 0.1 \\
        \midrule
        StreamMapNet~\cite{Yuan_2024_streammapnet} + HiVT~\cite{9878832} & 3.5E-4 & 32 & 0.1 \\
        StreamMapNet~\cite{Yuan_2024_streammapnet} + DenseTNT~\cite{Gu_Sun_Zhao_2021} & 1.0E-3 & 16 & 0.1 \\
        \bottomrule
    \end{tabular}
    }
    \label{tab:hyperparameters_trajectory}
\end{table}

\subsection{Quantitative Evaluation}
To evaluate the impact of the proposed uncertainties on downstream vehicle trajectory prediction, we conduct experiments comparing our method with previous uncertainty approaches across 8 model combinations. These combinations pair map information from 4 existing high-definition map estimation methods~\cite{MapTR, maptrv2,Yuan_2024_streammapnet} with 2 trajectory prediction methods~\cite{9878832, Gu_Sun_Zhao_2021}. 

\noindent\textbf{From the trajectory prediction aspects, we observe a consistent improvement in Table \ref{tbl:result}.} 
(1) Integration with MapTR: When incorporating MapTR for map estimation enhanced by our positional and semantic uncertainty, the DenseTNT trajectory prediction method demonstrates the most substantial improvements, with minADE, minFDE, and MR metrics enhanced by approximately 13\%, 5\%, and 2\%, respectively.
(2) Integration with MapTRv2: Despite MapTRv2 exhibiting superior performance over MapTR in HD map estimation tasks, its direct application in downstream trajectory prediction does not consistently yield improved outcomes and occasionally even results in slight performance reductions. However, incorporating our positional and semantic uncertainty, MapTRv2-based map predictions achieve performance gains comparable to those obtained with MapTR.
(3) Integration with MapTRv2-Centerline: Utilizing MapTRv2-centerline, which incorporates lane centerlines into map estimation, and applying our uncertainty estimations, both trajectory prediction methods attain their highest performance levels. Specifically, the HiVT method shows notable improvements, with minADE, minFDE, and MR metrics improved by 8\%, 10\%, and 22\%, respectively, over the baseline. The DenseTNT method, while less significantly enhanced, still achieves a 9\% improvement in minADE.
(4) Integration with StreamMapNet: In scenarios utilizing StreamMapNet for map construction complemented by our positional and semantic uncertainty estimations, both trajectory prediction methods again achieve optimal performance. Notably, the HiVT method demonstrates a particularly significant improvement, with the MR metric enhanced by 22\% compared to Gu~\etal~\cite{GuSongEtAl2024}.

\noindent\textbf{From the map aspects, the HiVT trajectory prediction model shows greater improvements.}
After applying our positional uncertainty and semantic uncertainty to all map methods, the improvement in MR is the most significant in HiVT, achieving an improvement of up to 22\%, indicating that by incorporating our proposed map uncertainty, the prediction model can effectively adjust its behavior to better match the actual trajectory. For DenseTNT, minADE shows the largest gain across the four map estimators, with up to a 13\% reduction. This suggests that our uncertainty module curbs large-displacement outliers and improves trajectory accuracy.
Overall, as shown in Table~\ref{tbl:result}, the predicted maps obtained using our positional uncertainty and semantic uncertainty achieve a significant performance improvement in downstream vehicle trajectory prediction compared to the baselines. 
\begin{figure*}[t]
    \centering
    \includegraphics[width=\linewidth]{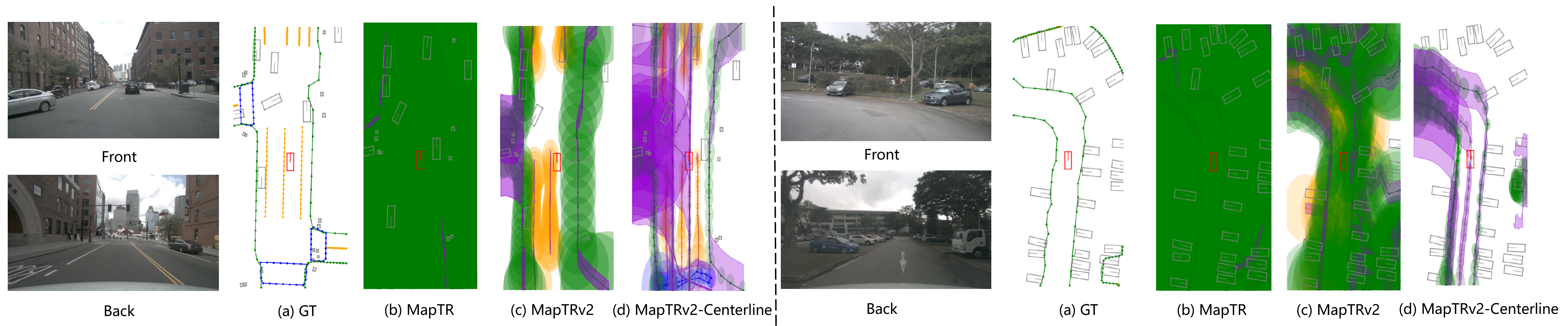}
    \caption{The left figure shows the effectiveness of our proposed method for estimating high-definition map positional uncertainty and semantic uncertainty in a normal road scenario in the test set. The right figure also shows the effectiveness of our proposed method for estimating high-definition map positional uncertainty and semantic uncertainty in test set scenarios involving curved roads and parking lots. \textcolor{ForestGreen}{Green} denotes road boundaries, \textcolor{blue}{blue} denotes pedestrian crossings, \textcolor{YellowOrange}{orange} denotes lane dividers, \textcolor{purple}{purple} indicates category semantic uncertainty, \textcolor{gray}{gray} denotes lane centerlines, the \textcolor{red}{red} vehicle denotes the ego vehicle, and the \textcolor{gray}{gray vehicles} denote other agents.
    }
    \label{fig:map1}
\end{figure*}
\subsection{Ablation Studies and Further Discussion}
In Table~\ref{tbl:ablation}, we study the impact of our proposed positional and semantic uncertainties with two different trajectory prediction methods, \ie, HiVT~\cite{9878832} and DenseTNT~\cite{Gu_Sun_Zhao_2021}.

\noindent\textbf{Effectiveness of Positional Uncertainty.} We first compare the effect of introducing only positional uncertainty of map elements against the baseline method. Based on the same trajectory prediction method, \ie, HiVT, all four trajectory prediction evaluation metrics show an improvement, with minADE increasing the most by 8\%. This indicates that the HiVT-based method is more sensitive to the accuracy of the positional information of map elements. In contrast, for DenseTNT, introducing only positional uncertainty to enhance map elements does not yield a significant improvement in trajectory prediction, and even leads to a decline in MR. This suggests that DenseTNT, utilizing GNN, is already capable of effectively leveraging positional relationships of map elements.

\noindent\textbf{Effectiveness of Semantic Uncertainty.} When introducing only semantic uncertainty to enhance map elements, the performance on HiVT improves more significantly compared to using positional uncertainty alone, particularly with a 9\% increase in minADE compared to the baseline. For DenseTNT, the introduction of semantic uncertainty yields substantial improvements, with minADE increasing by 10\% and minFDE by 7\%. This demonstrates that the accuracy of semantic information plays a crucial role in enhancing trajectory prediction in complex and occluded scenarios. Since there are inherent errors in map estimation compared to ground truth, incorporating uncertainty in category information can better assist the trajectory prediction model.

Overall, applying both positional and semantic uncertainty map information to the HiVT model results in more noticeable improvements than DenseTNT. Introducing positional and semantic uncertainty information into the HiVT trajectory prediction model consistently enhances predictions, with semantic uncertainty showing a greater impact. Notably, when both uncertainties are utilized together, the MR metric for HiVT improves significantly, whereas the improvement is minimal when using either one individually. This highlights that the proposed positional and semantic uncertainties are both indispensable and complementary.

\noindent\textbf{The impact of adding modules on model training and testing.} To obtain the final predicted trajectories, our approach adopts a two-stage training strategy. In the first stage, we employ four distinct HD map estimation models to integrate positional and semantic uncertainty estimations, yielding maps that comprehensively account for these uncertainties. After processing, these uncertainty-informed maps are input into the second stage, where two representative trajectory prediction models, Transformer-based and GNN-based, are utilized to produce the final trajectory predictions.
Using the MapTRv2~\cite{maptrv2} combined with HiVT~\cite{9878832} as an illustrative example, we evaluate the impact of our method on training and inference times using the NVIDIA A6000 GPU. In the first stage of HD map estimation, the baseline~\cite{GuSongEtAl2024} approach requires 1 day and 8 minutes to train, with approximately 40.64 M trainable parameters. After incorporating our uncertainty estimation technique, the training time slightly increases to 1 day and 2 hours, with the number of trainable parameters rising modestly to approximately 43.08 M. This demonstrates that our uncertainty estimation does not significantly increase either the training time or model complexity in the HD map estimation phase.
In the second stage, the baseline trajectory prediction method requires approximately 8 minutes and 50 seconds per epoch, with an estimated total parameter size of 17.43 M. When incorporating maps enhanced by our uncertainty information, training time increases marginally to 9 minutes per epoch, with nearly identical parameter size at approximately 17.436 M. Furthermore, the inference time remains identical in both scenarios. Thus, our dual-head uncertainty estimation module and its integration into the trajectory prediction stage introduce minimal overhead in terms of training time and model parameters while substantially enhancing overall trajectory prediction performance.

\noindent\textbf{Map Uncertainty Visualization.}
In Figure~\ref{fig:map1}, we illustrate the visualization effects of the two uncertainties introduced across the four map estimation methods. The top of the figure shows a scenario where tall buildings on both sides of the road obscure the intersection, and the presence of other vehicles and pedestrians results in incomplete information captured by the camera of the vehicle, leading to high uncertainty in the map model prediction. It can be observed that the MapTR model generates high levels of positional and semantic uncertainty, whereas MapTRv2 and MapTRv2-centerline exhibit lower uncertainty. However, the obscured intersection causes these models to produce higher positional and semantic uncertainty at the road junction. The bottom of the figure illustrates a parking lot environment, where road boundaries are unclear and there are no distinct driving lanes, with many surrounding vehicles obscuring the road conditions. Here, our positional and semantic uncertainties are particularly evident at the turns, reflecting the changes in the road under such conditions.

\begin{figure*}[t]
    \centering
    \includegraphics[width=\linewidth]{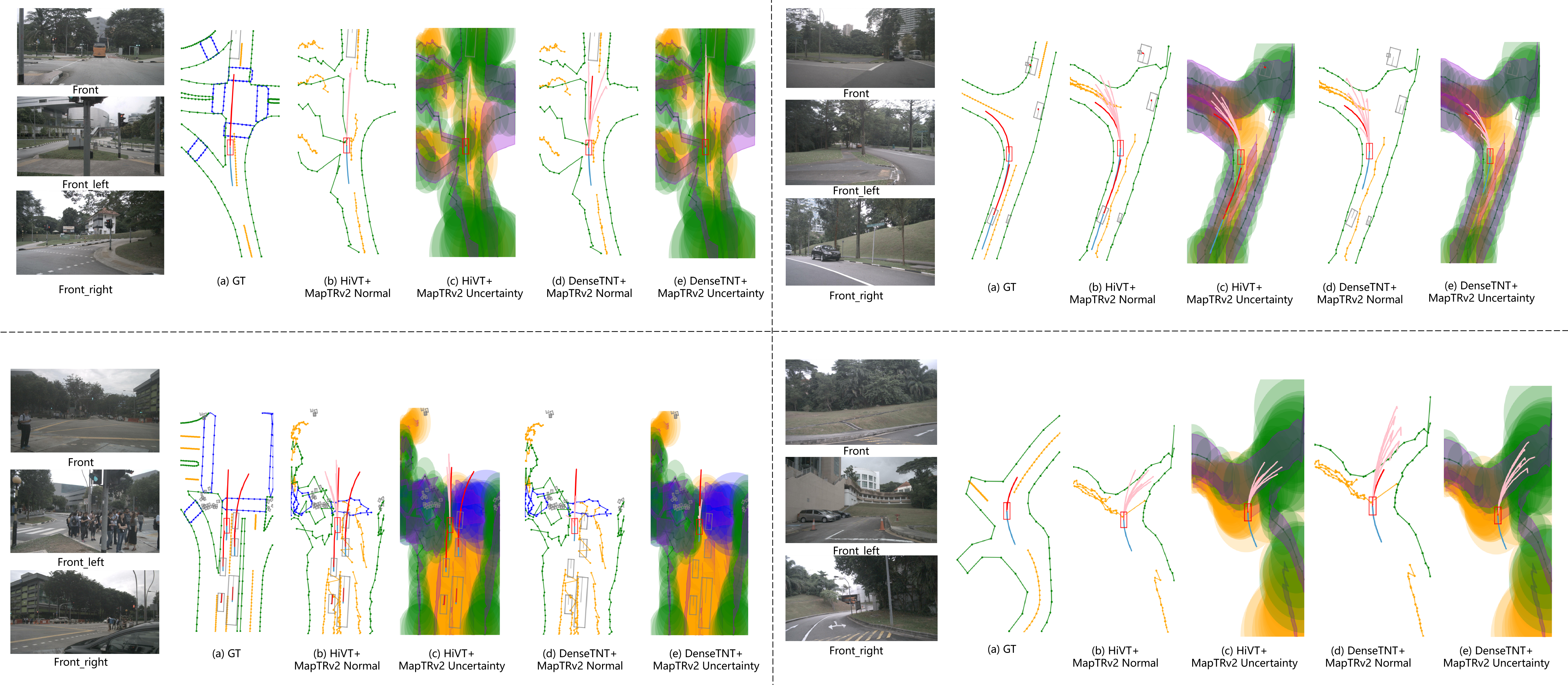}
    \caption{\textbf{Top Left:} At busy intersections with dense map elements, our proposed uncertainty information improves vehicle trajectory prediction. \textbf{Top Right:} When turning, the camera perspective often fails to capture all surrounding road conditions, potentially causing trajectory predictions to extend beyond the road boundaries. \textbf{Bottom Left:} In complex environments with numerous occlusions, both types of uncertainties in map prediction increase. \textbf{Bottom Right:} When lane information is unclear, the environment is open, and map estimation is poor, our uncertainty information helps maintain high accuracy in trajectory prediction despite incomplete map input.
    }
    \label{fig:vis1}
\end{figure*}

\noindent\textbf{Uncertainty-aware Trajectory Prediction Visualization.}
To better illustrate the improvement in map trajectory prediction brought by our proposed positional uncertainty and semantic uncertainty, we visualize the enhancement effects in some typical scenarios using our two types of uncertainty in Figure~\ref{fig:vis1}. For a clearer representation of how these uncertainties supplement map information, we choose MapTRv2 to generate visualization images with two trajectory prediction models.
\textbf{(1) Complex Urban Intersections.} As shown in Figure~\ref{fig:vis1} top left, we evaluate vehicle trajectory predictions at a complex intersection with additional turning lanes. The figure includes the ground truth of the map and vehicle trajectories. We observe that using HiVT and DenseTNT as inputs for the downstream trajectory prediction tasks, with the same map uncertainty, results in competitive prediction performance, reducing routing errors and effectively handling such multi-lane scenarios. Especially in the case of DenseTNT, the vehicle's predicted trajectory is noticeably closer to the ground truth due to the additional support from both types of map uncertainty.
\textbf{(2) Vehicle Turning Scenario.} In Figure~\ref{fig:vis1} top right, we show a scenario where the vehicle is about to turn, and the trajectory prediction model is prone to large errors due to unclear road boundaries and camera perspective issues. By incorporating our two types of uncertainty in the map information, we can clearly see that the vehicle trajectory in both methods is more reasonable, avoiding situations where the trajectory exceeds road boundaries when no uncertainty is introduced.
\textbf{(3) Traffic Situation with Significant Occlusion.} As shown in Figure~\ref{fig:vis1} bottom left, we present the improvement in model trajectory prediction in a complex traffic situation with significant occlusion and many pedestrians. When many pedestrians obscure the road information, our introduced uncertainties are reflected in darker colors, indicating the model uncertainty about both the positional and semantic information in these areas. Without such uncertainty assistance, the model predicted trajectory can be seen to deviate significantly, suggesting that the vehicle would drive toward the pedestrians. By incorporating both positional and semantic uncertainty, the model considers these uncertainties and predicts a more reasonable trajectory.
\textbf{(4) Unclear Map Environment.} In Figure~\ref{fig:vis1} bottom right, we illustrate a scenario where the road environment is relatively open, the road information is vague, and the existing map estimation models are unable to accurately estimate all map elements. By introducing the two types of uncertainty, \ie, positional and semantic uncertainty, we can supplement the map information and obtain more accurate model predictions. The figure shows that when there are fewer map elements without uncertainty supplementation, the vehicle trajectory tends to drift beyond the road boundaries. However, after introducing the uncertainty information, the situation is alleviated, allowing for reasonable trajectory prediction despite the lack of complete map element information.

\begin{figure}[t]
    \centering    \includegraphics[width=0.9\linewidth]{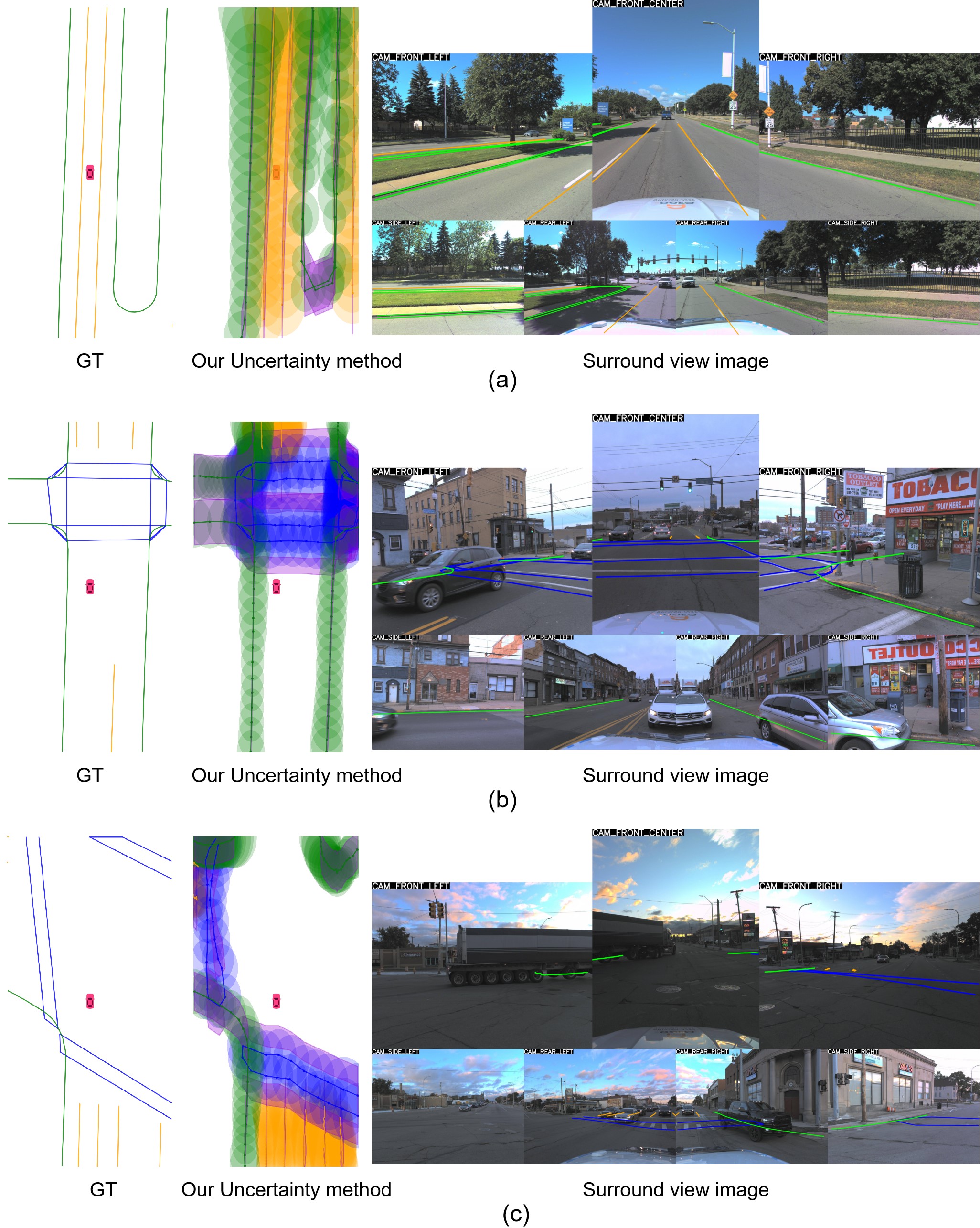}
    \caption{Map visualization of our uncertainty method in the Argoversev2 sensor dataset. In the figure, \textcolor{ForestGreen}{green} represents road boundaries, \textcolor{blue}{blue} represents pedestrian crossings, \textcolor{orange}{orange} represents lane dividers, \textcolor{purple}{purple} indicates category semantic uncertainty, \textcolor{red}{red} vehicle denotes the ego vehicle. 
    }
    \label{fig:argo}
\end{figure}

\noindent\textbf{Map Uncertainty Visualization in Argoversev2.}
In Figure~\ref{fig:argo}, we present the HD map visualization on the Argoverse2 sensor dataset after applying our proposed uncertainty estimation. In (a), the tree‐obscured rear corner shows elevated semantic and positional uncertainties, whereas other portions of the road exhibit consistently low semantic uncertainty. (b) and (c) capture more complex intersections, where occlusions, particularly by corner obstructions or turning vehicles (e.g., the oil tanker in (c)), raise both semantic and positional uncertainties. These uncertainties manifest as darker and broader regions around obscured road boundaries and pedestrian lines. In contrast, when both sides of the road boundary are unobstructed, the semantic uncertainty is consistently low, and the positional uncertainty, though still higher at intersections, remains comparatively moderate.

\section{Case Studies}
\begin{figure}[t]
    \centering
    \includegraphics[width=0.98\linewidth]{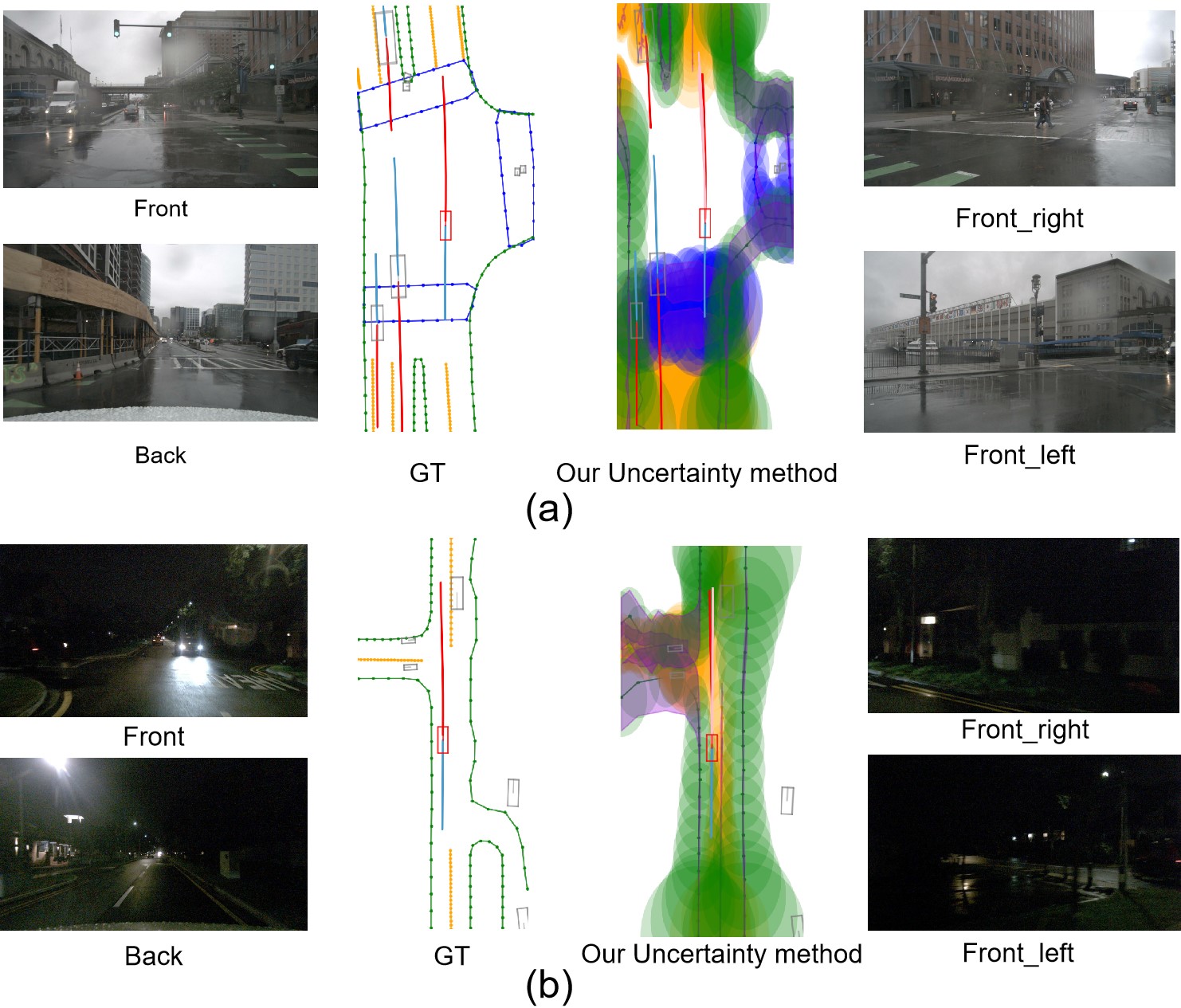}
    \caption{Visualization of our uncertainty estimation in challenging conditions: (a) rainy scenario with visibility distortions, and (b) nighttime scenario with limited visibility. Our method effectively highlights positional and label semantic uncertainties, estimates the map uncertainty information, and demonstrates robustness in diverse environments.}
    \label{fig:vis2}
\end{figure}

 \textbf{Challenging cases}. In Figure~\ref{fig:vis2}, we illustrate the effectiveness of our uncertainty estimation method under challenging conditions. (a) shows a rainy scene characterized by visual distortions such as overcast weather, road surface reflections from water, and raindrops. These factors degrade camera perception, increasing uncertainty in detecting road edges and lane markings. Our method accurately identifies and quantifies these uncertainties, allowing the vehicle to maintain a precise trajectory. Areas with obscured or blurred visibility, such as sidewalks blocked by vehicles and rain, exhibit larger and more prominent circles of positional and semantic uncertainty, demonstrating the robustness of our approach. (b) presents a nighttime scenario where poor visibility hampers image sensor accuracy, leading to higher uncertainty in road location estimations compared to daytime conditions. At an intersection obscured by trees on the left side of the vehicle, our method effectively highlights increased semantic uncertainty and positional uncertainty in both road and lane lines. Additionally, positional uncertainty is notably higher at the periphery of the field of view, consistent with real-world vehicle navigation requirements. These results validate the efficacy of our uncertainty estimation in diverse and demanding environments.

\begin{figure}[t]
    \centering
    \includegraphics[width=\linewidth]{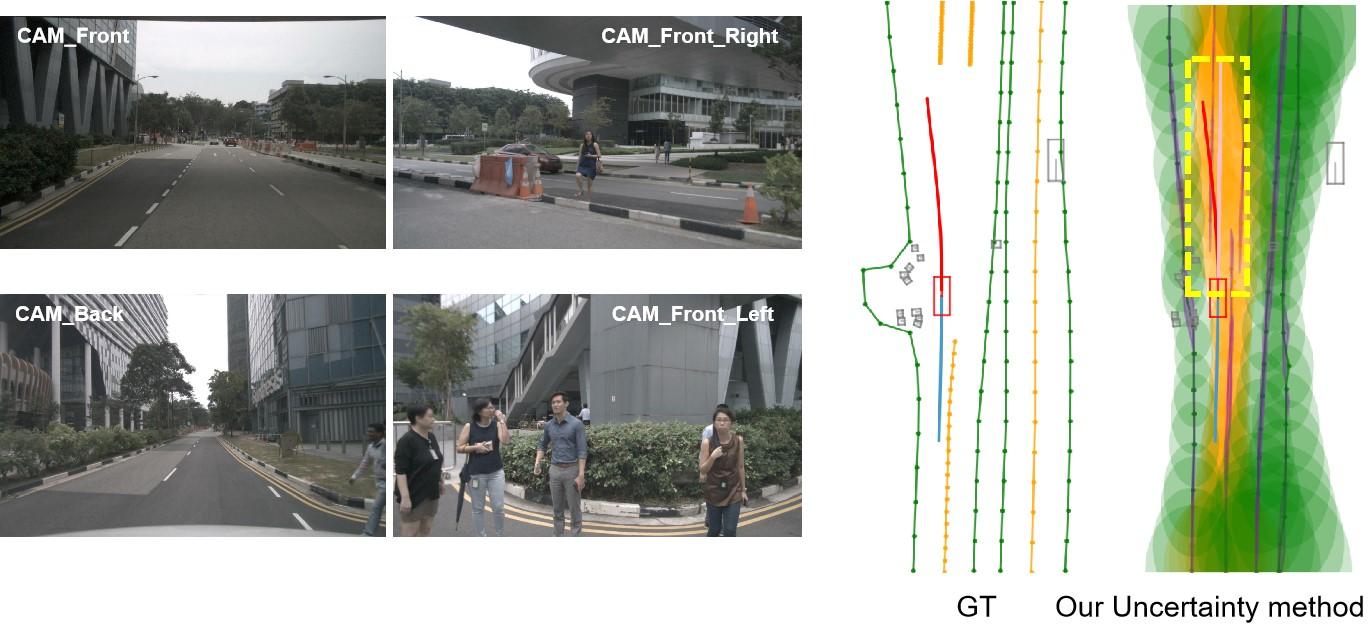}
    \caption{Visualization of a failure case. The yellow box in the figure illustrates the discrepancy between the ground truth and our predicted trajectory, highlighting a case where an abnormal ground truth deviation under simple, unobstructed road conditions leads to prediction errors. }
    \label{fig:vis3}
\end{figure}

 \noindent\textbf{Failure case}. We also identify a failure case for our uncertainty estimation framework, visualized in Figure~\ref{fig:vis3}. As highlighted by the yellow box, the ground truth trajectory (in red) shows a noticeable leftward deviation, while our predicted trajectory (in pink) expects the vehicle to continue straight along the current road. Given that the vehicle is on a flat road without obstructions, complex conditions, or intersection turns, it is reasonable to predict that the vehicle would continue straight. However, the ground truth trajectory from the nuScenes dataset, despite being labeled as "straight," shows a slight leftward drift, causing the prediction discrepancy. Thus, our uncertainty estimation method may be affected when ground-truth trajectories deviate unexpectedly from normal driving expectations, even under simple road conditions.

\section{Conclusion}
In this work, we propose a universal uncertainty estimation framework for vectorized HD maps, addressing map noise propagation in trajectory prediction via an auxiliary prediction head that jointly regresses positional and semantic uncertainties. By integrating this framework into the state-of-the-art online map estimators, including MapTR, MapTRv2 variants, and StreamMapNet, we generate uncertainty-aware map elements. These enhanced representations are evaluated through downstream trajectory predictors, \eg, DenseTNT and HiVT, demonstrating maximum improvements of 8\% minADE, 10\% minFDE, and 22\% MR reduction. Our results validate that explicit uncertainty decomposition effectively mitigates map noise interference in prediction tasks.
\bibliographystyle{IEEEtran}
\bibliography{egbib}
\begin{IEEEbiography}[{\includegraphics[width=1in,height=1.2in, clip,keepaspectratio]{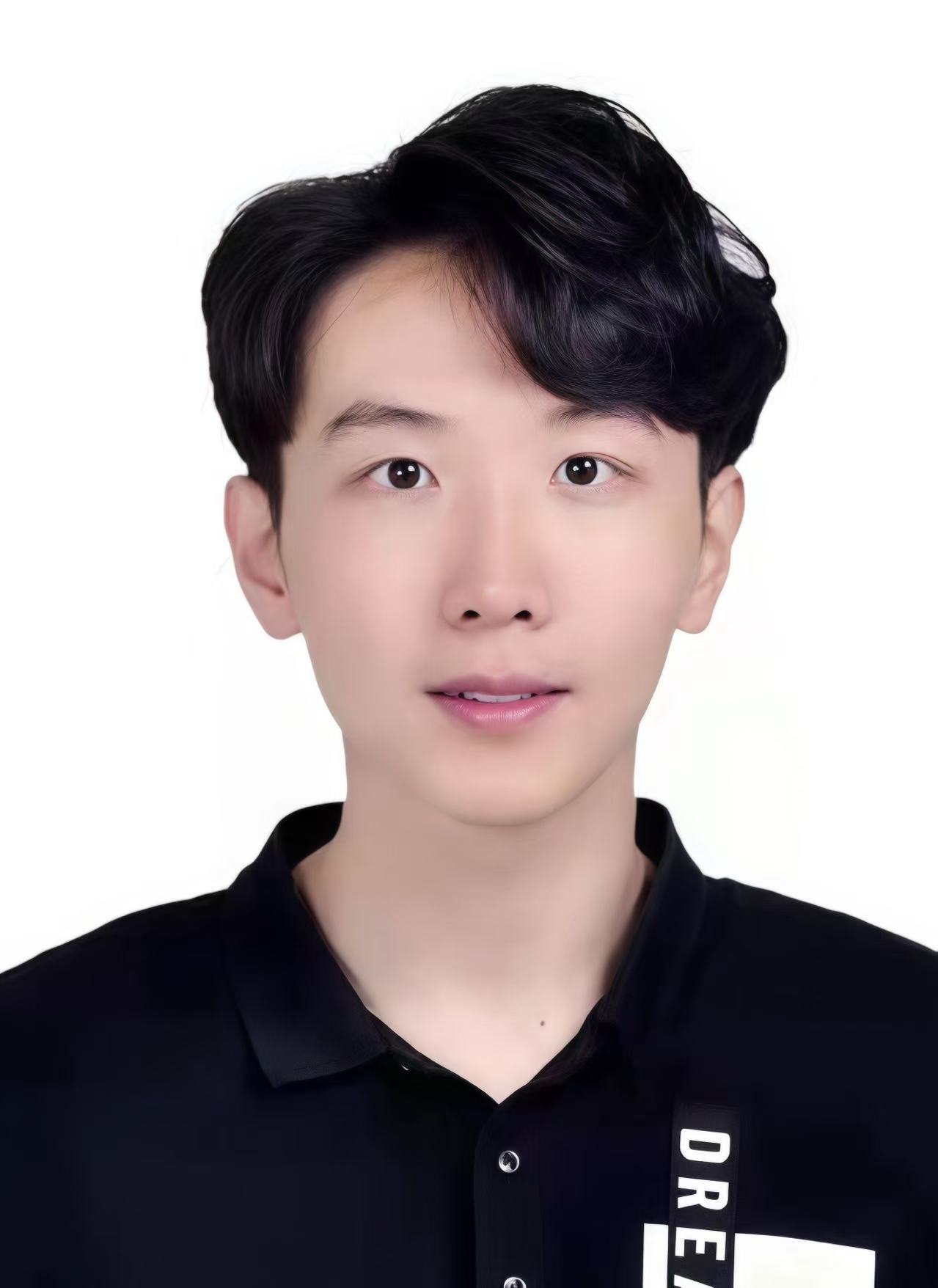}}]{Jintao Sun} received the M.S. degree from The George Washington University, America, in 2021 and the B.E. degree from the Harbin Institute of Technology, China, in 2019. He is currently working toward the Ph.D. degree with the School of Computer Science and Technology, Beijing Institute of Technology, China. His current research interests include image retrieval, vision and language models and autonomous driving. 
\end{IEEEbiography}

\begin{IEEEbiography}[{\includegraphics[width=1in,height=1.25in, clip,keepaspectratio]{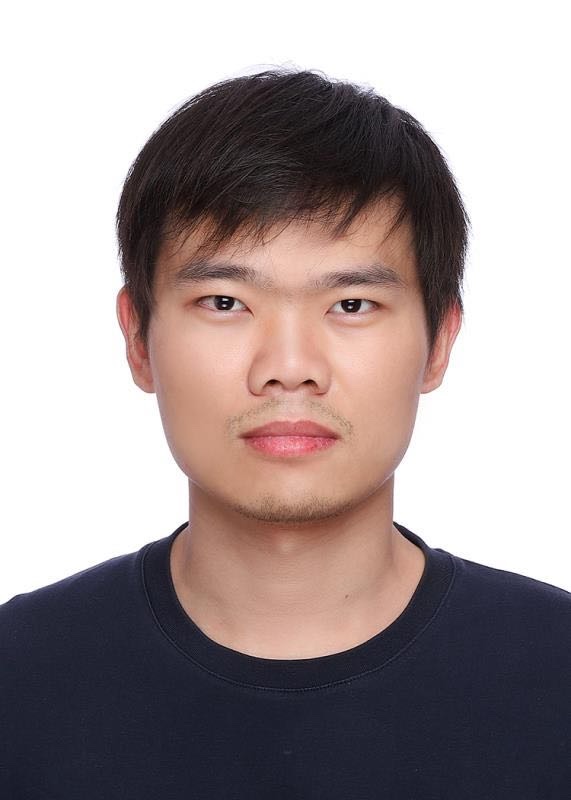}}]{Hu Zhang} is a CERC Research Fellow in CSIRO DATA61, Australia. He received the Ph.D. degree from the University of Technology Sydney in 2022 and the B.S. degree from the University of Science and Technology of China in 2017. His research interests include 3D computer vision in Autonomous Driving, 3D reconstruction, imbalanced data learning.
\end{IEEEbiography}

\begin{IEEEbiography}[{\includegraphics[width=1in,height=1.2in,clip,keepaspectratio]{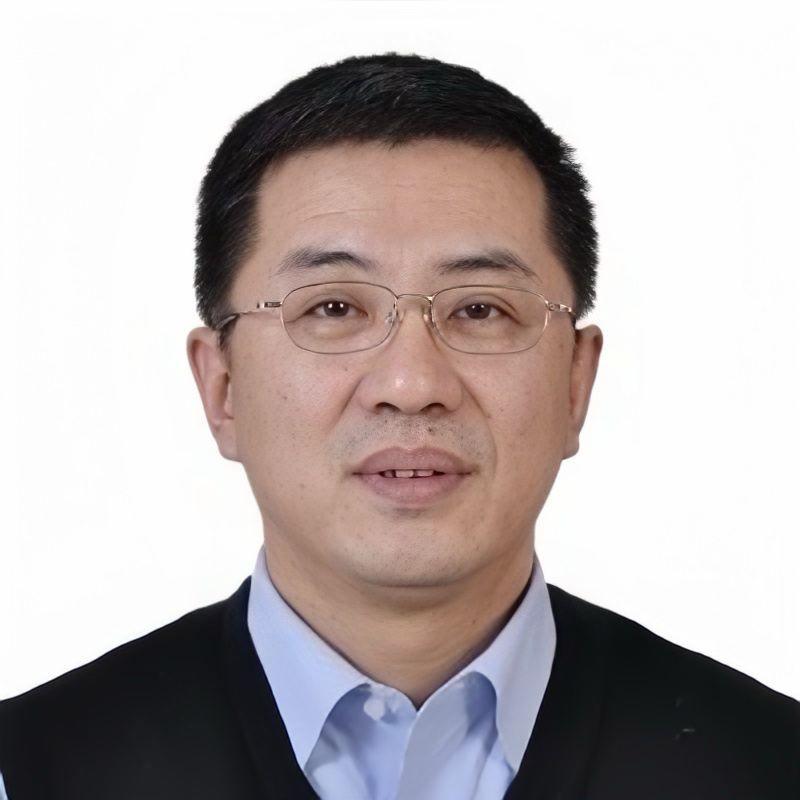}}]{Gangyi Ding} received the B.E. degree from Peking
University, Beijing, China, in 1988 and the Ph.D.
degree from the Beijing Institute of Technology, Beijing, in 1993. He is currently a Professor with the School of Computer Science and Technology, Beijing
Institute of Technology. In 1993, he joined the faculty,
Beijing Institute of Technology. His research interests
include computer simulation, software engineering,
and digital performance.
\end{IEEEbiography}

\begin{IEEEbiography}[{\includegraphics[width=1in,height=1.25in, clip,keepaspectratio]{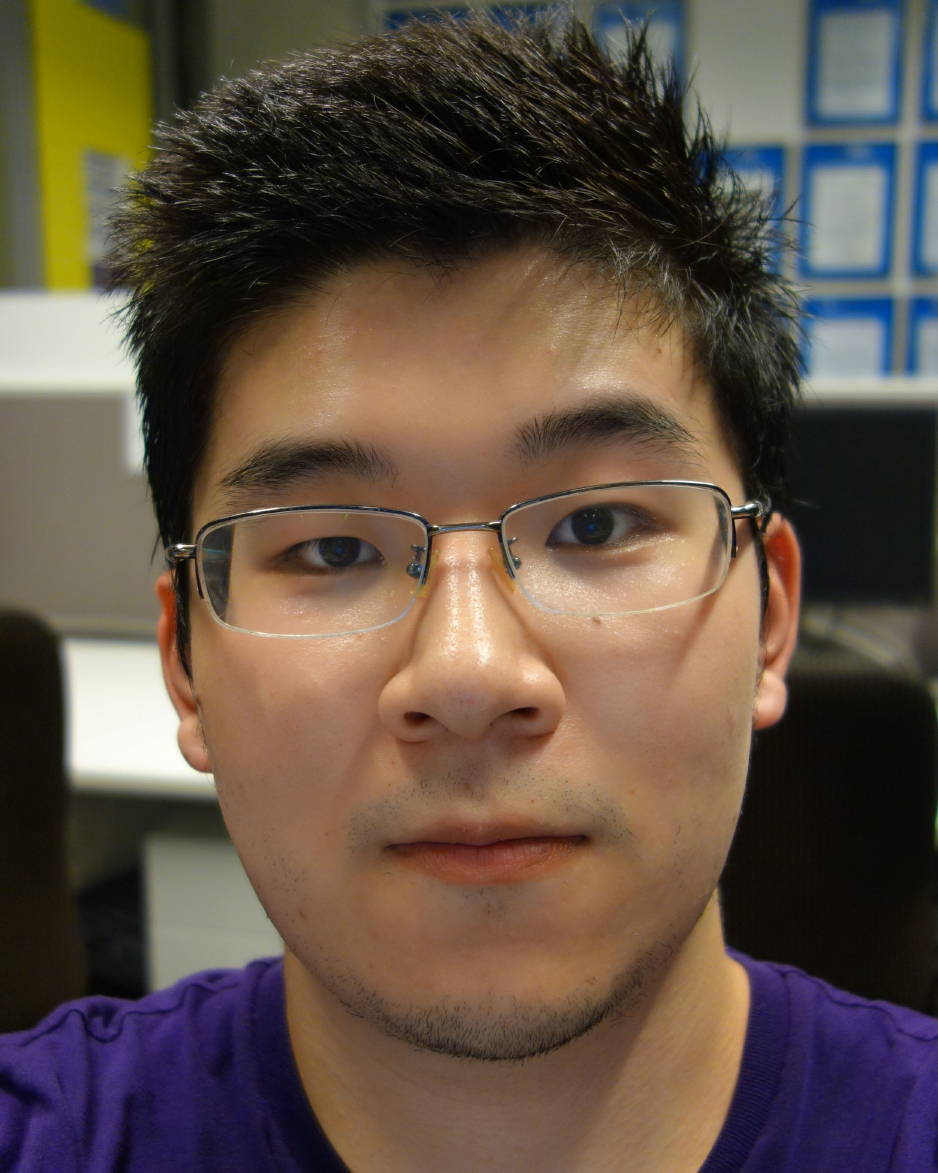}}]{Zhedong Zheng} is an Assistant Professor with the University of Macau. He received the Ph.D. degree from the University of Technology Sydney in 2021 and the B.S. degree from Fudan University in 2016. He was a postdoctoral research fellow at the School of Computing, National University of Singapore. He received the IEEE Circuits and Systems Society Outstanding Young Author Award of 2021. His research interests include AIGC, Data-centric AI, and Spatial Intelligence. He actively serves the academic community, acting as a Senior PC for IJCAI and AAAI, an Area Chair for ACM MM'24, ACM MM'25 and ICASSP'25, and the Publication Chair for ACM MM'25 and AVSS'25.
\end{IEEEbiography}



\end{document}